\documentclass[11pt,a4paper]{article}
\usepackage[hyperref]{acl2019}
\usepackage{times}
\usepackage{latexsym}
\usepackage{CJKutf8}

\usepackage{url}

\usepackage{bbm}
\usepackage{graphicx}
\usepackage{booktabs} 
\usepackage{makecell}

\graphicspath{ {./images/} }

\aclfinalcopy 
\def\aclpaperid{295} 

\title{Robust Neural Machine Translation with Joint Textual and Phonetic Embedding}
\author{Hairong Liu$^1$ \,
Mingbo Ma$^{1,3}$ \,
Liang Huang$^{1,3}$ \,
Hao Xiong$^2$ \,
Zhongjun He$^2$
\\[0.1cm]
  $^{1}$Baidu Research, Sunnyvale, CA, USA
  \\ 
  $^2$Baidu, Inc., Beijing, China  
\\
  $^3$Oregon State University, Corvallis, OR, USA
\\
  {\tt \small \{liuhairong, mingboma, lianghuang, xionghao05, hezhongjun\}@baidu.com }
}

\date{}

\begin{document}
\maketitle
\begin{CJK}{UTF8}{gbsn}
\begin{abstract}
  Neural machine translation (NMT) is notoriously sensitive to noises, but noises are almost inevitable in practice. One special kind of noise is the \emph{homophone noise}, where words are replaced by other words with similar pronunciations.\footnote{In this paper,  the word ``homophone" is loosely used to represent characters or words with similar pronunciations.} We propose to improve the robustness of NMT to homophone noises by 1) jointly embedding both textual and phonetic information of source sentences,
  and 2) augmenting the training dataset with homophone noises.
  Interestingly, to achieve better translation quality and more robustness, we found that most (though not all) weights should be put on the phonetic rather than textual information.  Experiments show that our method not only significantly improves the robustness of NMT to homophone noises,  but also surprisingly improves the translation quality on some clean test sets.
\end{abstract}

\section{Introduction}
Recently we witnessed tremendous progresses in the field of neural machine translation (NMT) \cite{sutskever2014sequence, cho2014learning, bahdanau2014neural, luong2015effective, gehring2017convolutional}, especially the birth of \emph{transformer network} \cite{vaswani2017attention}.

Despite tremendous success, NMT models are very sensitive to the noises in input sentences \cite{belinkov2017synthetic}. The causes of such vulnerability are multifold, and some of them are: 1) neural networks are inherently sensitive to noises, such as adversarial examples \cite{goodfellow2014generative, szegedy2013intriguing}, 2) every input word can affect every output word generated by the decoder due to the global effects of attention,  and 3) all NMT models have an input embedding layer, which is sensitive to noises in the input sentences.
\begin{table*}[h]
\begin{center}
 \begin{tabular}{ll} \toprule
   Clean Input & 目前已发现\underline{有}109人死亡, 另有57人获救  \\
   Output of Transformer & at present, 109 people have been found dead and 57 have been rescued \\
   \midrule
   Noisy Input & 目前已发现\underline{又}109人死亡, 另有57人获救 \\
   Output of Transformer& the hpv has been found dead so far and 57 have been saved \\
   Output of Our Method & so far, 109 people have been found dead and 57 others have been rescued  \\
   \bottomrule
    \end{tabular}
 \caption{The translation results on Mandarin sentences without and with homophone noises.  The word `有' (y\v{o}u, ``have'') in clean input is replaced by one of its homophone, `又' (y\`{o}u, ``again''), to form a noisy input. This seemingly minor change completely fools the Transformer to generate something irrelvant (``hpv''). 
   Our method, by contrast, is very robust to homophone noises thanks to the usage of phonetic information.}
    \label{tab:sensitive_example}
  \end{center}
\end{table*}

In this paper, we focus on \emph{homophone noise}, where words are replaced by other words with similar pronunciations, which is common in real-world systems.
One example is speech translation \cite{ruiz2017assessing,ruiz2015phonetically,ma+:2018}, where an ASR system may output correct or almost correct phoneme sequences, but transcribe some words into their homophones.
Another example is pronunciation-based input systems for non-phonetic writing systems such as Pinyin for Chinese or Katakana/Hiragana for Japanese.
It is very common for a user to accidentally choose a homophone instead of the correct word.
Existing NMT systems are very sensitive to homophone noises, and Table \ref{tab:sensitive_example} illustrates such an example. The transformer model can correctly translate the clean input sentence; however, when one Mandarin character, `有'(y\v{o}u), is replaced by one of its homophones, `又'(y\`{o}u), the transformer generates a strange and irrelevant translation. The method proposed in this paper can generate correct results under such kind of noises, since it uses both textual and phonetic information.

Since words are discrete signals, to feed them into a neural network, a common practice is to encode them into real-valued vectors through \emph{embedding}. However, the output of the embedding layer is very sensitive to noises in the input sentences. This is because when a word $a$ is replaced by another word $b$ with different meanings, the embedding vector of $b$ may be very different from the embedding vector of $a$, thus results in dramatic changes. To make things worse, the input embedding layer is usually the first layer of the network, and errors from this layer will propagate and be amplified in the following layers, leading to more severe errors. For homophone noises, since correct phonetic information exists, we can make use of it to make the output of the embedding layer much more robust.  

In this paper, we propose to improve the robustness of NMT models to homophone noises by jointly embedding both textual and phonetic information. In our approach, both words and their corresponding pronunciations are embedded and then combined to feed into a neural network. 
This approach has the following advantages:
\begin{enumerate}
\item It is a simple but general approach, and easy to implement. 
\item It can dramatically improve the robustness of NMT models to homophone noises.
\item It also improves translation quality on clean test sets.
\end{enumerate}

To further improve the robustness of NMT models to homophone noises, we use \emph{data augmentation}  to expand the training datasets, by randomly adding homophone noises. The experimental results clearly show that data augmentation improves the robustness of NMT models\footnote{See more information and our code at \url{https://phoneticmt.github.io/}}.

\section{Joint Embedding}


For a word $a$ in the source language, suppose its pronunciation can be expressed by a sequence of \emph{pronunciation units}, such as phonemes or syllables, denoted by $\Psi(a)=\{s_1, \ldots, s_l\}$. Note that we use the term ``word" loosely here,  and in fact $a$ may be a word or a subword, or even a character. 

We embed both pronunciation units and words, and both of them are learnt from scratch. For a pronunciation unit $s$, its embedding is denoted by $\pi(s)$, and for a word $a$, its embedding is denoted by $\pi(a)$. For a pair of a word $a$ and its pronunciation sequence $\psi(a)=\{s_1, \ldots, s_l\}$, we have $l + 1$ embedding vectors, that is, $\pi(a), \pi(s_1), ..., \pi(s_l)$. To get a fixed length vector representation, we first merge $\pi(s_1), ..., \pi(s_l)$ into a single vector by averaging, denoted by $\pi(\psi(a))$,\footnote{We tried other approaches, such as using an LSTM network to merge them; however,  we did not see obvious improvements in translation quality.} then combine the word embedding and $\pi(\psi(a))$ as follows:
\begin{equation}
\pi([a, \psi(a)]) = (1-\beta)*\pi(a) + \beta*\pi(\psi(a))
\end{equation}
where $\beta$ is a parameter. When $\beta=0$, only textual embedding is used; while when $\beta=1$,  only phonetical embedding is used . The best balance, as demonstrated by our experiments, is a very large $\beta$ close to but not $1$. 

\section{Experiments}
\subsection{Models}
In our experiments, we use Transformer as baseline. Specifically, we use the PyTorch version (PyTorch 0.4.0) of OpenNMT.  All models are trained with $8$ GPUs, and the values of important parameters are: 6 layers, 8 heads attention,  2048 neurons in feed-forward layer, and 512 neurons in other layers, dropout rate is $0.1$, label smoothing rate is $0.1$, Adam optimizer, learning rate is $2$ with NOAM decay. 

\subsection{Translation Tasks}
We evaluate our method on the translation task of Mandarin to English, and reported the $4$-gram BLEU score \cite{papineni2002bleu} as calculated by \emph{multi-bleu.perl}.

Pinyin is used as pronunciation units \cite{du2017pinyin, yang2018improved}, and there are $404$ types of pinyin syllables in total \footnote{For simplicity reasons, tone information is discarded.}. A large Mandarin lexicon is used. For words or subwords not in the lexicon, if all of their characters have pinyins, the concatenation of these characters's pinyins are used as the pinyin of the whole words or subwords. Note that when there are multiple pronunciations, we just randomly pick one in both training and testing. For symbols or entries without pronunciation, we use a special pronunciation unit, $\langle unk \rangle$, to represent them.

\subsection{Translation Results}
For the dataset, we use an extended NIST corpus which consists of $2$M sentence pairs with about $51$M Mandarin words and $62$M English words, respectively. We apply byte-pair encodings (BPE) \cite{sennrich2016neural} on both Mandarin and English sides to reduce the vocabulary size down to $18$K and $10$K, respectively. Sentences longer than $256$ subwords or words are excluded.

\begin{figure}[h]
\includegraphics[scale=0.35]{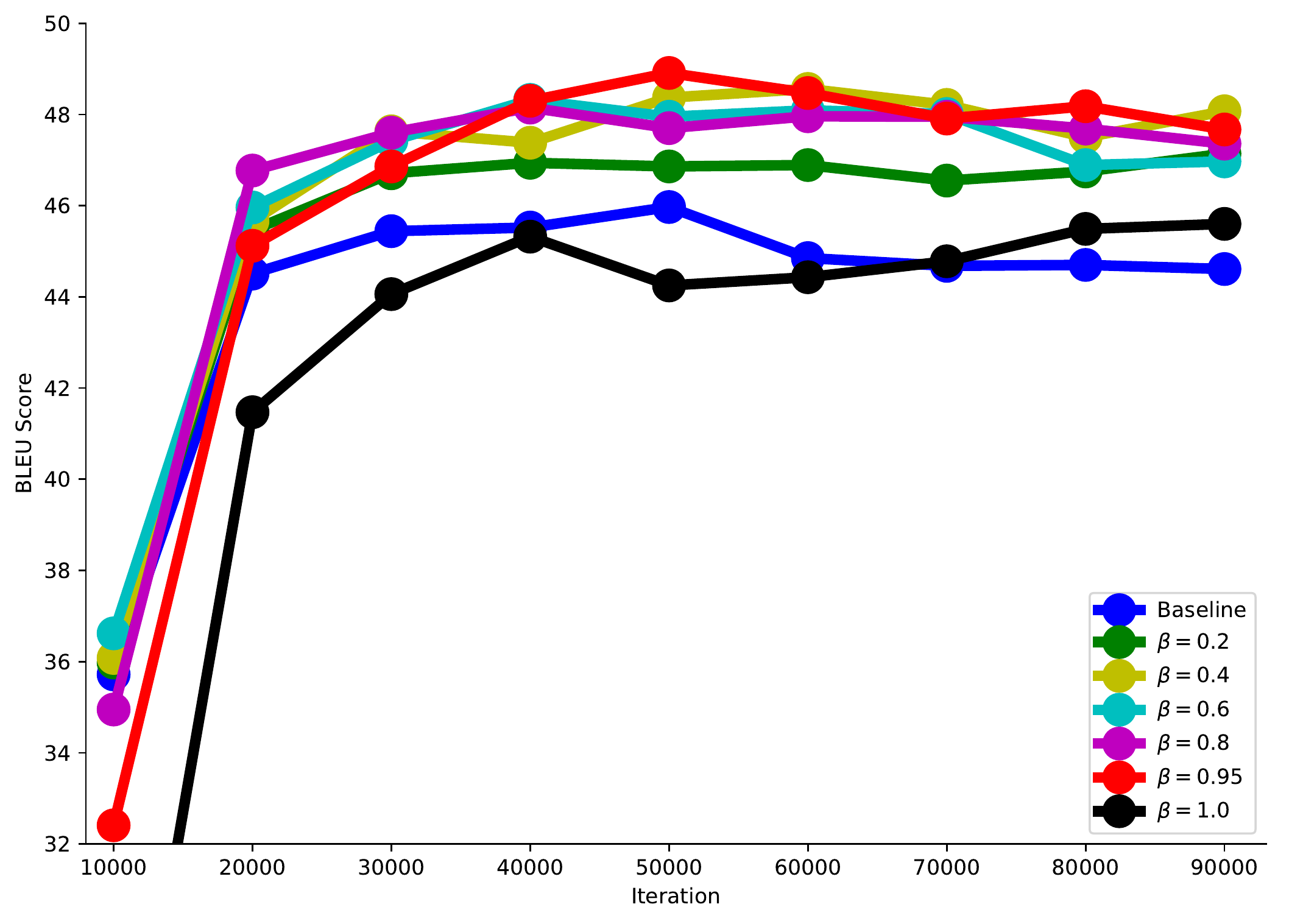}
\caption{BLEU scores on the dev set for the baseline model (Transformer-base) and our models with different $\beta$. The $x$-axis is the number of iterations and the $y$-axis in the case-insensitive BLEU scores on multiple references. }
\label{fig:iter_alpha}
\end{figure}
In Figure \ref{fig:iter_alpha}, we compare the performances, measured by BLEU scores to multiple references, of the baseline model and our models with $\beta=0.2, 0.4, 0.6, 0.8, 0.95, 1.0$, respectively. We report the results every $10000$ iterations from iteration $10000$ to iteration $90000$. Note that our model is almost exactly the same as baseline model, with only different source embeddings. In theory, when $\beta=0$, our model is identical to baseline model. However, in practice, there is a slight difference: when $\beta=0$, the embedding parameters are still there, which will affect the optimization procedure even no gradients flow back to these parameters. When $\beta=1$, only phonetic information is used. There are some interesting observations from Figure \ref{fig:iter_alpha}. First, combing textual and phonetic information improves the performance of translation. Compared with baseline, when $\beta=0.2$, the BLEU scores improves $1-2$ points, and when $\beta=0.4, 0.6, 0.8, 0.95$, the BLEU scores improves $2-3$ points. Second, 
the phonetic information plays a very important role in translation. Even when $\beta=0.95$, that is, the weight of phonetic embedding is $0.95$ and the weight of word embedding is only $0.05$, the performance is still very good. In fact, our best BLEU score ($48.91$), is achieved when $\beta=0.95$. However, word embedding is still important. In fact, when we use only phonetic information (when $\beta=1.0$), the performance become worse, almost the same as baseline (only using textual information). Our human only needs phonetic information to communicate with each other, this is probably because we have better ability to understand context than machines, thus do not need the help of textual information.

\begin{table*}[h]
 \begin{center}
\begin{tabular}{l|c|cccc} \toprule
Models & \makecell{NIST06 \\ (Dev Set)} & NIST02 & NIST03 & NIST04 & NIST08 \\ \midrule
  Transformer-base & $45.97$ & $47.40$ & $46.01$ & $47.25$ & $41.71$ \\
  $\beta=0.2$ & $47.14$ & $48.63$ & $47.82$ & $48.63$ & $43.77$ \\
  $\beta=0.4$ & $48.56$ & $49.41$ & $48.73$& $50.53$ & $\mathbf{45.16}$ \\
  $\beta=0.6$ & $48.32$ & $48.83$ & $48.82$ & $49.86$ & $44.17$ \\
  $\beta=0.8$ & $48.15$ & $\mathbf{49.42}$ & $49.44$ & $49.98$ & $44.86$ \\
  $\beta=0.95$ & $\mathbf{48.91}$ & $49.33$ & $\mathbf{50.46}$ & $\mathbf{50.57}$ & $44.83$ \\
  $\beta=1.0$ & $45.6$ & $47.04$ & $46.42$ & $47.65$ & $40.27$ \\
\bottomrule
\end{tabular}
  \caption{Translation results on NIST Mandarin-English test sets}
  \label{tab:alpha_5test}
 \end{center}
\end{table*}

Table \ref{tab:alpha_5test} reports results on the baseline model and our models under different $\beta$s. NIST $06$ is used as dev set to select the best models, and NIST $2002$, $2003$, $2004$, $2005$ and $2008$ datasets are used as test sets. There are some interesting observations. First, combing textual and phonetic information improves the performance of translation. This seems to be surprising since no additional information is provided. Although the real reason is unknown, we suspect that it is because some kind of regularization effects from phonetic embeddings. Second, 
the phonetic information plays a very important role in translation. Even when $\beta=0.95$, that is, most weights are put on phonetic embedding, the performance is still very good. In fact, our best BLEU score ($48.91$), is achieved when $\beta=0.95$. However, word embedding is still important. In fact, when we use only phonetic information ($\beta=1.0$), the performance degrades, almost the same as baseline (only using textual information).

%


\begin{figure}[h]
\includegraphics[scale=0.40]{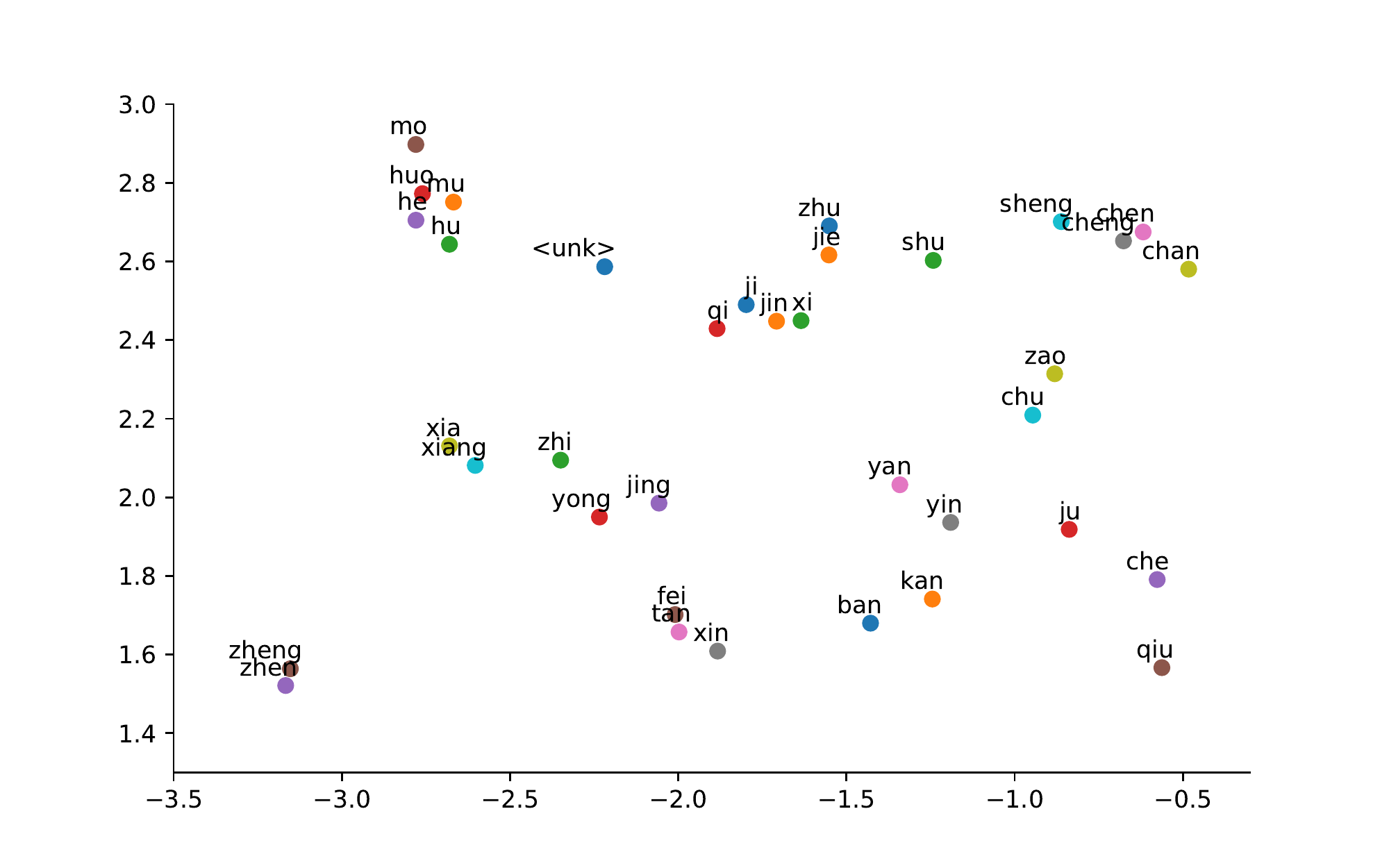}
\caption{Visualization of a small region in the embedding space. Note that pinyins with similar pronunciations are close in the embedding space.}
\label{fig:phoneme_emb}
\end{figure}
To understand why phonetic information helps the translation, it is helpful to visualize the embedding of pronunciation units. We projects the whole Pinyin embedding space into a $2$-dimensional space using t-SNE technique \cite{maaten2008visualizing}, and illustrate a small region of it in Figure \ref{fig:phoneme_emb}. An intriguing property of the embedding is that pinyins with similar pronunciations are close to each other, such as \textit{zhen} and \textit{zheng},  \textit{ji} and \textit{qi},  \textit{mu} and \textit{hu}.  This is very helpful since in Mandarin, two characters with similar pronunciations will either $1)$ be represented by the same pinyin or $2)$ be represented by two pinyins with similar pronunciations. 

Homophones are very common in Mandarin. In our training dataset, about $55\%$ Mandarin words have homophones.  To test the robustness of NMT models to homophone noises, we created two noisy test sets, namely, NoisySet$1$, and NoisySet$2$, based on NIST$06$ Mandarin-English test set. The creation procedure is as follows: for each source sentence in NIST$06$, we scan it from left to right, and if a word has homophones, it will be replaced by one of its homophones by a certain probability ($10\%$  for NoisySet$1$ and $20\%$ for NoisySet$2$).

\begin{figure}[ht]
\includegraphics[scale=0.36]{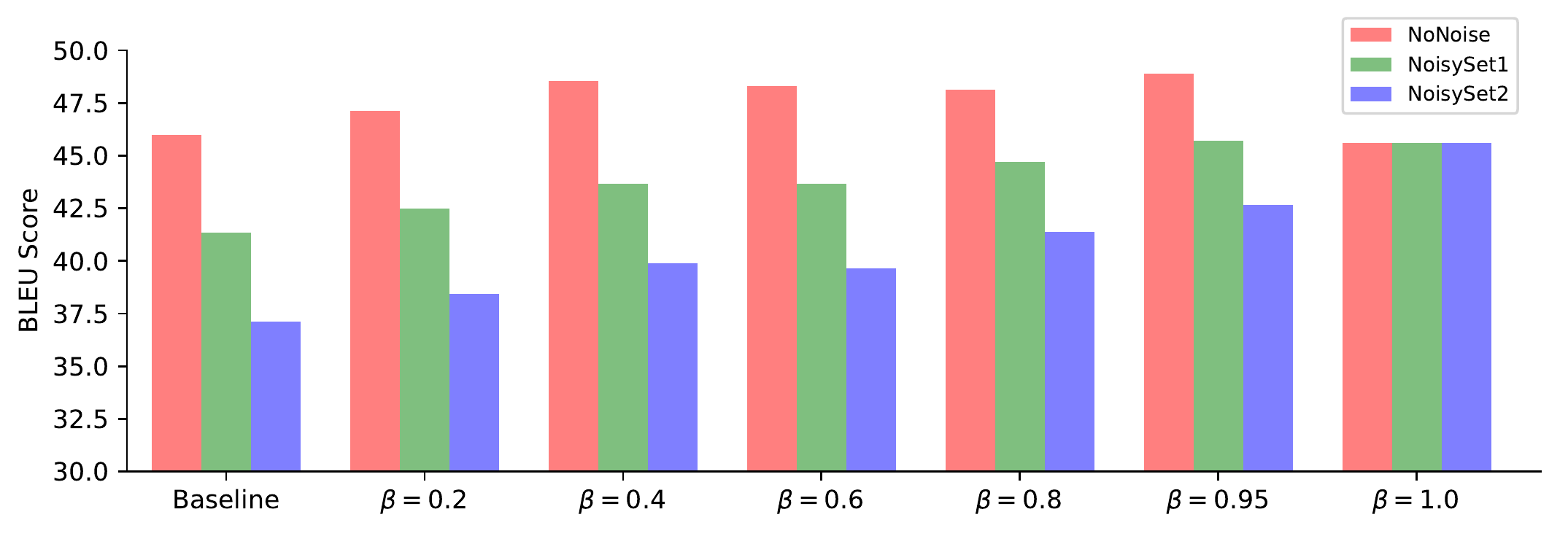}
\caption{BLEU scores on dataset without and with homophone noises. On both noisy test sets, as more weight are put on phonetic embedding, that is, as $\beta$ grows, the translation quality improves.}
\label{fig:noise_cmp}
\end{figure}

In Figure \ref{fig:noise_cmp}, we compare the performance of the baseline model and our models with $\beta=0.2, 0.4, 0.6, 0.8, 0.95, 1.0$, respectively, on NIST$06$ test set and the two created noisy sets. The models are chosen based on their performance (BLEU scores) on NIST$06$ test set. As Figure \ref{fig:noise_cmp} shows, as $\beta$ grows, which means that more weights are put on phonetic information, the performances on both noisy test sets almost steadily improve. When $\beta=1.0$, as expected, homophone noises will not affect the results since the model is trained solely based on phonetic information.
However, this is not our best choice since the performance on the clean test set gets much worse. In fact, from the perspective of robustness to homophone noises, the best choice of $\beta$ is still a value smaller but close to $1$, which mainly focuses on phonetic information but still utilizes some textual information.

\begin{table*}[h]
\begin{center}
 \begin{tabular}{ll} \toprule
   Clean Input & 古巴是第一个\underline{与}(y\v{u})新中国建交的拉美国家   \\
   Output of Transformer & cuba was the first latin american country to \\
                                       & establish diplomatic relations with new china \\
   Noisy Input & 古巴是第一个\underline{于}(y\'{u})新中国建交的拉美国家  \\
   Output of Transformer & cuba was the first latin american country to discovering the establishment of \\ 
                                        & diplomatic relations between china and new Zealand \\
   Output of Our Method & cuba is the first latin american country to \\
                                       & establish diplomatic relations with new china  \\
   \midrule  
   Clean Input & 他认为, 格方\underline{对}(du\`{i})俄方的指责是荒谬的   \\
   Output of Transformer & he believes that georgia's accusation against russia is absurd \\
   Noisy Input & 他认为, 格方\underline{憝}(du\`{i})俄方的指责是荒谬的  \\
   Output of Transformer & he believes that the accusations by the russian side villains are absurd  \\
   Output of Our Method &  he maintained that georgia's accusation against russia is absurd \\
   \bottomrule
    \end{tabular}
    \caption{Two examples of homophone noises on source sentences.  The underscored Mandarin characters are homophones, and their corresponding Pinyin pronunciations are in the parentheses. Note that textual-only embedding is very sensitive to homophone noises, thus generates weird outputs. However, when jointly embedding both textual and phonetic information in source sentences, the model is very robust.}
    \label{tab:noisy_examples}
  \end{center}
\end{table*}

\begin{table*}[ht]
\begin{tabular}{lcccccc} \toprule
Models & \multicolumn{3}{c}{Before Augmentation} &  \multicolumn{3}{c}{After Augmentation} \\
\cmidrule(lr){2-4} \cmidrule(lr){5-7} \\
&NIST$06$ & NoisySet$1$ & NoisySet$2$ & NIST$06$ & NoisySet$1$ & NoisySet$2$ \\
  Transformer-base & $45.97$ & $41.33$ & $37.11$ & $43.94$ & $42.61$ & $41.33$ \\
  $\beta=0.95$ & $\mathbf{48.91}$ & $\mathbf{45.71}$ & $\mathbf{42.66}$ & $\mathbf{48.06}$ & $\mathbf{47.37}$ & $\mathbf{46.47}$ \\
\bottomrule
\end{tabular}
 \caption{Comparison of models trained with and without data augmentation. }
\label{tab:data_aug}
\end{table*}

Table \ref{tab:noisy_examples} demonstrate the effects of homophone noises on two sentences. The baseline model can translate both sentences correctly; however, when only one word (preposition) is replaced by one of its homophones, the baseline model generates incorrect, redundant and strange translations. This shows the vulnerability of the baseline model. Note that since the replaced words are prepositions,  the meaning of the noisy source sentences are still very clear, and it does not affect our human's understanding at all. For our method, we use the model with $\beta=0.95$, and it generates reasonable translations. 




To further improve the robustness of NMT models, we augment the training dataset by randomly picking training pairs from training datasets, and revising the source sentences by randomly replacing some words with their homophones. We add $40\%$ noisy sentence pairs on the original $2$M sentence pairs in the training set, resulting in a training dataset with about $2.8$M sentence pairs.

In Table \ref{tab:data_aug}, we report the performance of baseline model and our model with $\beta=0.95$, with and without data augmentation. Not surprisingly, data augmentation significantly improves the robustness of NMT models to homophone noises. However, the noises in training data seem to hurt the performance of the baseline model (from $45.97$ to $43.94$), and its effect on our model seems to be much smaller, probably because our model mainly uses the phonetic information.



\section{Related Work}
\citet{formiga2012dealing}  proposed to use a character-level translator to deal with misspelled words in the input sentences, but in general their method cannot deal with homophone noises effectively. \citet{cheng2018towards}  proposed to use adversarial stability training to improve the robustness of NMT systems, but their method does not specifically target homophone noises and do not use phonetic information. The effects of ASR errors on machine translation have been extensively analyzed \cite{ruiz2017assessing, ruiz2015phonetically}. In a parallel work, \citet{li2018improving} also proposed to utilize both textual and phonetic information to improve the robustness of NMT systems, but their method is different with ours in how textual and phonetic information are combined.

\section{Conclusion}
In this paper, we propose to use both textual and phonetic information in NMT by combining them in the input embedding layer of neural networks. Such combination not only makes NMT models much more robust to homophone noises, but also improves their performance on clean datasets. Our experimental results clearly show that both textual and phonetical information are important, and the best choice is to rely mostly on phonetic information. We also augment the training dataset by adding homophone noises, and our experiments demonstrate that this is very useful in improving the robustness of NMT models to homophone noises.

\clearpage

\bibliography{phonemeMT}
\bibliographystyle{acl_natbib}

\end{CJK}

\end{document}